\documentclass[10pt]{article} 
\usepackage[preprint]{tmlr}

\usepackage{amsmath,amsfonts,bm}

\def\eqref#1{equation~\ref{#1}}

\def\1{\bm{1}}

\DeclareMathAlphabet{\mathsfit}{\encodingdefault}{\sfdefault}{m}{sl}
\SetMathAlphabet{\mathsfit}{bold}{\encodingdefault}{\sfdefault}{bx}{n}

\usepackage{hyperref}
\usepackage{url}
\usepackage{graphicx}
\usepackage{booktabs}
\usepackage{amsmath}
\usepackage{amssymb}

\title{Grokking Is Conditional and Fragile: A Fully-Tractable, Multi-Seed Study at 12K Parameters}

\author{\name Yoshiyuki Ootani \email info@ootanl.com \\
      \addr Independent Researcher}

\def\month{07}  
\def\year{2026} 
\def\openreview{\url{https://openreview.net/forum?id=XXXX}} 

\begin{document}

\maketitle

\begin{abstract}
Grokking, the delayed onset of generalization long after a network has fit its training set, is
usually studied in models large enough that the learned algorithm is only partially legible and
where a single training run is taken as evidence. We instead study an $\approx$11{,}856-parameter
Llama-style transformer (Glimmer-1-Base) on modular arithmetic, a regime small enough to enumerate
and read its weights, attention, and full input-output map directly, and we measure grokking as a
multi-seed \emph{rate} rather than a single outcome. Six findings emerge. (i) The grokking
transition is gated by training-set coverage, and the coverage threshold tracks
the output cardinality (the modulus) more than composition structure, an ordering that holds above
the transition and across a ten-fold change in domain size. (ii) Weight decay reproduces the Omnigrok
inverted-U at 12K parameters, grok-rate rising from 20\% to 90\% then collapsing to 0\%, a positive
control on the rate measurement. (iii) Grokking sits on a
numerical knife-edge: two distinct perturbations of the floating-point environment, the CPU thread
count (a pure change of reduction order) and CPU-versus-GPU execution (a device change that subsumes
it), each flip a minority of seeds (49/300 and 19/100 at the 0.70 threshold) without a detectable
shift in the aggregate rate (well powered: exact McNemar $p\ge0.39$; paired-difference 95\%
confidence intervals within $\pm10$ points). (iv) The mechanistic read-out is mixed:
generalizing solutions have a more periodic output map (a partly definitional consistency check),
while the genuinely independent result is a \emph{negative} one, namely that the dim-16 model does
not form the textbook Fourier embedding circle. (v) Task decomposition helps not by enabling an
otherwise impossible computation but by converting a coverage-starved sparse task into densely
coverable sub-tasks; at a matched data budget a two-specialist pipeline groks where the monolith
cannot (10/10 versus 0/10), while a same-budget scratchpad monolith carrying the identical
decomposed supervision still fails (0/10), isolating coverage rather than supervision density as the
driver. (vi) Methodologically, multi-seed
control with a fixed numerical environment overturns three dramatic single-run narratives in our
own data, a hard task ``wall'', a ``thread count flips grokking'' effect, and a ``GPU suppresses
grokking'' effect, each of which proves to be a seed confound. In this fully-tractable regime,
grokking is best understood as a conditional and fragile phase transition, and multi-seed numerical
control is a precondition for any claim about it.
\end{abstract}

\section{Introduction}
\label{sec:intro}

A neural network can fit a small algorithmic dataset to perfect training accuracy and continue,
for thousands of further steps, to generalize no better than chance, and then, abruptly,
generalize almost perfectly. This delayed generalization, \emph{grokking} \citep{power2022grokking},
has become a central object of study for the science of deep learning because it isolates the
question of \emph{when} a network converts memorization into an algorithm. Most of what we know
about grokking comes either from mechanistic case studies of a single trained network
\citep{nanda2023progress} or from phenomenological accounts in models with millions of parameters
\citep{liu2022omnigrok,prieto2025edge}. Both settings share two limitations. The learned
computation is only partially legible, so claims about mechanism rest on probing rather than on a
complete reading of the weights; and grokking is reported from individual runs, even though it is
known to be seed-sensitive.

We take the opposite tack on both counts. We study Glimmer-1-Base \citep{glimmer1base2026}, a published
$\approx$11{,}856-parameter Llama-style transformer (hidden size 16, two layers, four attention
heads with a single key/value head, vocabulary 512), pretrained only on 500K tokens of FineWeb-Edu
and performing at chance on every standard benchmark. At this size the model is not a useful
language model, but it is something more valuable for our purpose: it is \emph{fully tractable}.
Its digit embeddings, attention pattern, and output logits over the entire input domain can be
enumerated and read directly. Smallness here is the experiment's resolution, not a defect we
tolerate: only a model this small can be read end to end, retrained cheaply enough to report
grok-\emph{rates} instead of single runs, and pinned tightly enough to hold its entire numerical
environment fixed -- control that confounds erase as parameter count grows. Tractability is thus
\emph{enabling} rather than a result in itself: it powers the mechanistic read-outs of
Section~\ref{sec:mechanism} and a clean negative, though not the complete causal circuit. We pair
this tractability with a methodological commitment: we never
report grokking from a single run. Every claim is a grok-\emph{rate} over ten or more seeds,
measured inside a fixed numerical environment whose thread count and device are themselves held
constant.

This combination matters, because the regime actively invites false stories. In the course of
this study our own single-seed observations suggested three clean, publishable narratives -- a hard
capability ``wall'', a ``thread count flips grokking'' effect, and a ``GPU suppresses grokking''
effect -- and multi-seed control refuted all three as seed confounds
(detailed in Section~\ref{sec:discussion}). We treat these refutations as a primary result: in this regime, the unit of evidence must be the multi-seed rate
under a pinned numerical environment.

Our contributions are:
\begin{enumerate}
\item \textbf{A coverage--cardinality pattern.} Above the transition, the grokking coverage
threshold tracks the output cardinality (the modulus $M$) more than the composition structure of the
task, an ordering that holds across a ten-fold change in input-domain size (Section~\ref{sec:coverage}).
\item \textbf{An Omnigrok reproduction at 12K parameters.} Weight decay produces the inverted-U in
grok-rate predicted by the loss-landscape account of grokking, validating the rate measurement as
a positive control (Section~\ref{sec:wd}).
\item \textbf{A replicated numerical knife-edge.} Two distinct perturbations of the floating-point
environment, reduction order (CPU thread count) and execution device (which subsumes it), each flip
a minority of same-seed outcomes (49/300 and 19/100 at the 0.70 threshold, TF32 disabled) with no aggregate bias---a well-powered null (exact McNemar $p=0.39$/$1.0$; Newcombe 95\% CIs within $\pm10$ points), a fragility axis distinct from the
softmax-collapse instability of \citet{prieto2025edge} (Section~\ref{sec:knifeedge}).
\item \textbf{A mechanistic read-out whose informative result is a negative one}: the dim-16 model
does \emph{not} form the textbook Fourier embedding circle of \citet{nanda2023progress}. The
positive correlate, that generalizing solutions have a more periodic output map, is a partly
definitional consistency check rather than an independent probe (Section~\ref{sec:mechanism}).
\item \textbf{Decomposition as data efficiency.} At a matched data budget, a pipeline of sub-task
specialists groks a composite task where the monolith cannot, by converting sparse coverage into
dense coverage; a same-budget scratchpad monolith carrying the identical decomposed supervision
still fails (0/10), isolating coverage rather than supervision as the driver, and the mechanism is
memorization of dense sub-domains (Section~\ref{sec:decomp}).
\item \textbf{A methodological contribution}, demonstrated end-to-end: the refutation of three
single-seed narratives, and a corresponding control protocol for grokking studies
(Sections~\ref{sec:knifeedge} and~\ref{sec:discussion}).
\end{enumerate}

\section{Related Work}
\label{sec:related}

\textbf{Grokking.} \citet{power2022grokking} introduced grokking on small algorithmic datasets,
showing generalization can improve far past the point of overfitting; they already observed that
how much a network generalizes depends on the \emph{fraction} of data used for training, a
dependence our coverage analysis (Section~\ref{sec:coverage}) makes quantitative. Subsequent work
has sought the mechanism. \citet{varma2023circuit} explain grokking through circuit efficiency,
positing a critical dataset size beyond which a generalizing circuit becomes more efficient than
memorization; this is the mechanistic counterpart of the coverage threshold we measure.
\citet{kumar2024grokking} recast the same delay as a lazy-to-rich training transition.
\citet{thilak2022slingshot} instead tie grokking to an optimizer-driven ``slingshot'' instability,
one of several accounts that locate grokking near an optimization edge. \citet{liu2022omnigrok},
in \emph{Omnigrok}, attribute grokking to a mismatch
between the training- and test-loss curves as functions of weight norm (the ``LU'' mechanism) and
show that weight decay, data size, and representation quality all modulate it. We reproduce the weight-decay dependence in a model two to three
orders of magnitude smaller (Section~\ref{sec:wd}) and use it as a positive control. A
complementary account, \citet{liu2022understanding}, casts grokking on a phase diagram whose
training-set-size axis selects between comprehension, grokking, memorization, and confusion -- again
the coverage axis of Section~\ref{sec:coverage}.

\textbf{Mechanistic interpretability of modular arithmetic.} \citet{nanda2023progress} fully
reverse-engineer a one-layer transformer trained on modular addition and find it implements
addition as rotation on a circle via discrete Fourier components, splitting training into
memorization, circuit formation, and cleanup, an early form of hidden progress
\citep{barak2022hidden}. This Fourier-multiplication picture is the
reference mechanism for our read-out in Section~\ref{sec:mechanism}. We find a weaker version of
it: generalizing solutions are periodic in their input-output map, but our dim-16 model does not
lay its digit embeddings on a clean Fourier circle, so we report periodicity at the level of the
logit map rather than the embedding. That one modular map admits more than one mechanism is documented: \citet{zhong2023clock} find
trained networks split between a ``clock'' (Fourier-circle) and a qualitatively different ``pizza''
algorithm, and \citet{chughtai2023toy} generalize this to arbitrary group operations;
\citet{gromov2023grokking} derives periodic (Fourier) feature-map weights in closed form, extended to
modular \emph{polynomials} by \citet{doshi2024polynomials} -- the multi-term setting our composite
tasks instantiate. All show periodic computation without embeddings on a clean circle, our negative. Our logit-level read-out is deliberately agnostic to which
algorithm a given seed implements.
Grokking has also been framed as competition between a memorizing dense and a generalizing sparse
subnetwork \citep{merrill2023two}, matching the grokker/memorizer split our read-out separates.

\textbf{Numerical effects in grokking.} Closest to our Section~\ref{sec:knifeedge} is
\citet{prieto2025edge}, who show that without regularization grokking tasks drive the softmax into
a floating-point failure they call Softmax Collapse, which \emph{prevents} grokking, and who fix
it with a stabilized activation. Their effect systematically suppresses generalization. Ours is
different in kind: a perturbation of the
reduction order or execution device that flips individual seeds without a detectable change in the
average grok-rate. Both point to grokking optimization operating near a numerical edge.

\textbf{Reproducibility and seed sensitivity.} That deep-learning results vary across seeds is
widely known, and accounting for this variance is itself a studied problem
\citep{bouthillier2021variance}; a single seed can be a lucky or unlucky outlier
\citep{picard2021torch}, the precise hazard our multi-seed-rate protocol is built around. Most
directly, \citet{summers2021nondeterminism} show that different sources of training
nondeterminism---seed, cuDNN, reduction order---produce statistically indistinguishable variance
with chaotic per-run sensitivity; we specialize this to grokking, where it surfaces as discrete
grok/no-grok flips of individual seeds. A related
caution is that a sharp ``emergent'' transition can be
an artifact of a discontinuous metric rather than a property of the model
\citep{schaeffer2023mirage}; we address this head-on by verifying that our grok-rate story is
invariant to the grok threshold (Section~\ref{sec:discussion}). Our contribution is to make the
seed, and the numerical environment that interacts with it, a controlled, first-class variable in
grokking specifically, and to show concretely how single-seed grokking claims mislead.

\textbf{Mixture-of-experts and decomposition.} Sparsely-gated mixtures of experts route inputs to
specialized sub-networks for capacity at fixed compute \citep{shazeer2017moe,fedus2022switch}. Our
decomposition result (Section~\ref{sec:decomp}) is related but distinct: the specialists are
trained on \emph{different sub-tasks} rather than routed within one task, and the benefit we
identify is a coverage/data-efficiency effect specific to the grokking regime, not a capacity or
throughput effect.

\section{Model and Methodology}
\label{sec:method}

\textbf{Model.} Glimmer-1-Base is a Llama-style decoder: hidden size 16, two layers, four query
heads with one key/value head (grouped-query attention), SiLU MLP of width 24, RMSNorm, rotary
position embeddings, tied input/output embeddings, vocabulary 512, context 512, float32, for
11{,}856 trainable parameters in total. It is byte-level BPE tokenized; the digits 0--9 are single
tokens. The released checkpoint is pretrained on 500K tokens of FineWeb-Edu and scores at chance
on standard reasoning and knowledge benchmarks, and generates incoherent text. We use it not as a
language model but as a tractable substrate for fine-tuning.

\textbf{Tasks.} We fine-tune on modular-arithmetic maps written as text, for example
\texttt{7+5=}~$\rightarrow$~\texttt{2} for $(a{+}b)\bmod 10$. We keep the modulus $M \leq 10$ so
that every answer and every intermediate result is a single digit (a single token), which makes
teacher-forced and free-generation accuracy coincide and keeps the read-outs clean. Tasks range
from single binary operations ($(a{+}b)\bmod M$, $(a\cdot b)\bmod M$) to four-input composites
($(a\cdot b + c\cdot d)\bmod M$, $((a{+}b)\cdot(c{+}d))\bmod M$).

\textbf{Training and evaluation.} Unless stated otherwise we use AdamW (learning rate
$3\times10^{-3}$, minibatch size 16, weight decay 0.01) for up to 1{,}200--1{,}500 epochs (full
passes over the training set), on CPU in float32. We define a run as having \emph{grokked} if its best held-out accuracy
reaches a threshold $\tau=0.70$, far above the entropy chance level $1/M$ (for the composite tasks,
whose answer marginals are mildly non-uniform, the majority-class baseline is somewhat higher than
$1/M$ but still well below $\tau$). We write $n$ for the number of training examples and define the
\emph{coverage} as $n/|\text{domain}|$, where the domain has $10^2$ tuples for binary operations,
$10^3$ for three-input and $10^4$ for four-input composites. We evaluate on a held-out split of the
input domain; for composite tasks the held-out set is a fixed 1{,}000-tuple sample drawn with a
frozen seed so that coverage and difficulty are constant across conditions.

\begin{figure}[t]
\begin{center}
\includegraphics[width=0.60\linewidth]{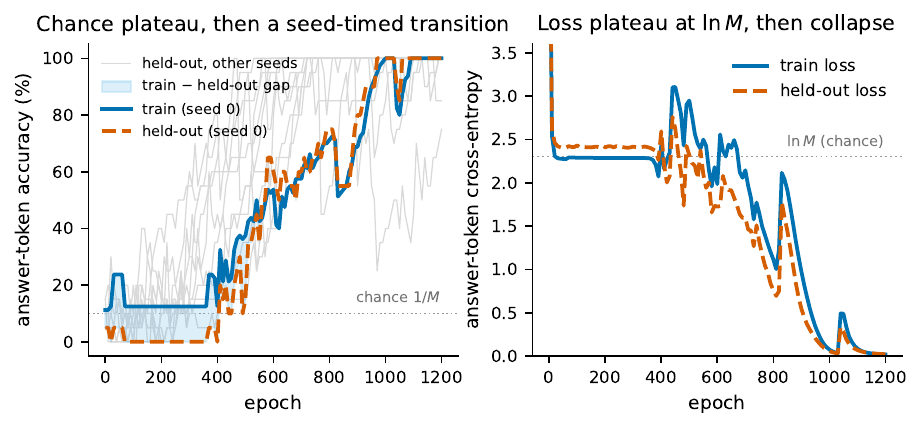}
\end{center}
\caption{Grokking dynamics for $(a{+}b)\bmod 10$ at weight decay 0.1 (all ten seeds grok). Left:
answer-token accuracy for a representative seed (bold) over the other nine seeds' held-out curves
(grey); held-out accuracy holds near chance, then transitions sharply at a seed-dependent epoch
($\approx$320--760). Right: train and held-out cross-entropy, plateaued at the chance value
$\ln M\approx2.30$ before collapsing. The brief shaded train$-$held-out gap shows memorization
precedes generalization only narrowly at this scale.}
\label{fig:trajectory}
\end{figure}

\textbf{What the best-accuracy summary collapses.} Defining grokking by best held-out accuracy
reduces a trajectory to a scalar; Figure~\ref{fig:trajectory} shows what that scalar summarizes. In
the grok-friendly regime the held-out loss holds at the chance value $\ln M$ for a seed-dependent
few hundred epochs and then collapses, the transition epoch ranging from $\approx$320 to 760 across
ten seeds -- the seed lottery in the time domain that the grok-rate aggregates over
\citep{notsawo2023predicting}. Within the
transition, training accuracy leads held-out by at most a few tenths, not by a long memorization
interval, so reducing each run to its best held-out
accuracy discards little.

\textbf{The multi-seed, fixed-environment protocol.} The central methodological commitment is that
no capability claim rests on a single run. Each condition is run over ten or more seeds and
reported as a grok-rate (and a mean best accuracy with its spread). Because grokking in this
regime is also sensitive to the floating-point environment (Section~\ref{sec:knifeedge}), we pin
the thread count and the device per study and report them. When we compare two conditions we pair
by seed so that the data split, the batch order, and the (deterministically loaded, pretrained)
initial weights are identical and only the variable under test differs. A grok-rate over ten seeds
carries a Wilson 95\% half-width of roughly 15--30 percentage points, so we treat rate differences
below about 30 points at ten seeds as within noise and rely on the 30-seed grids and paired tests
for the tightest comparisons.

\textbf{Why CPU.} At 11{,}856 parameters the arithmetic is trivial and kernel-launch overhead
dominates, so CPU is faster than a high-end GPU (67~s versus 162~s for a representative run; 301~s
versus 349~s in the paired comparison of Section~\ref{sec:knifeedge}). We therefore run all sweeps
on CPU and treat the GPU only as a second numerical environment for the device control -- for models
this size, the GPU offers no speed advantage, only a different numerical basis.

\paragraph{Use of generative AI.} In preparing this study the author used generative AI tools---AI coding/research agents and large language models---to help implement and run the experiments, analyse the results, and draft and edit the manuscript. All experimental results and claims were reviewed and verified by the author, who takes full responsibility for them.

\section{Results}
\label{sec:results}

\subsection{Coverage-gated grokking and the cardinality regularity}
\label{sec:coverage}

We first ask what controls whether a composite task groks at all. Fixing the held-out set and
varying only the fraction of the input domain seen during training (the \emph{coverage}), we find
a sharp, almost all-or-nothing transition: below a threshold coverage the grok-rate is near zero
and individual runs sit at chance-level held-out accuracy; above it the grok-rate jumps to one and
runs reach near-perfect accuracy. There is little intermediate behavior per seed.

\begin{figure}[t]
\begin{center}
\includegraphics[width=0.46\linewidth]{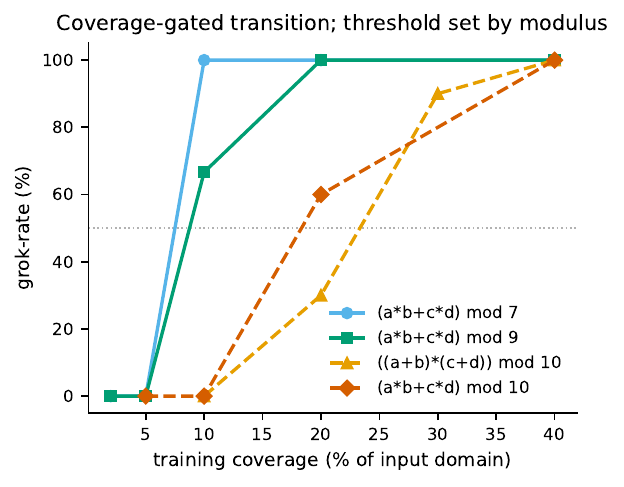}
\end{center}
\caption{Grok-rate versus training coverage for four composite tasks. Lower-cardinality moduli
($M=7,9$) cross over near 5--10\% coverage; both mod-10 tasks, despite different composition
structure, share the later 20--40\% threshold, showing the threshold is set by output cardinality,
not by operator structure.}
\label{fig:coverage}
\end{figure}

The threshold depends on the task (Table~\ref{tab:coverage}, Figure~\ref{fig:coverage}) and rises
with the modulus $M$ (the output cardinality): the lower-cardinality mod-7 and mod-9 composites
cross over near 5--10\% coverage, the mod-10 composite near 20--40\%. Composition structure matters
less. The sum-of-products $(a\cdot b + c\cdot d)\bmod 10$ and the structurally different
product-of-sums $((a{+}b)\cdot(c{+}d))\bmod 10$ cross over in the same $\sim$20--40\% band, so the
dominant determinant of the threshold is \emph{how many distinct answers the model must separate},
with \emph{which computation} produces them a secondary factor.

\begin{table}[t]
\caption{Grok-rate versus training coverage. The 0$\rightarrow$full transition is sharp; its
location tracks the modulus $M$ (output cardinality) and is shared by the two mod-10 tasks despite
their different structure. Cells use 3--5 seeds (mod-7/9 and product-of-sums: 3; mod-10: 5), so the
mod-9 and product-of-sums thresholds are provisional; the denser sweeps in Table~\ref{tab:struct}
and Figure~\ref{fig:cardinality} use ten seeds.}
\label{tab:coverage}
\begin{center}
\begin{tabular}{lccccc}
\toprule
task & 5\% & 10\% & 20\% & 40\% & threshold \\
\midrule
$(a\cdot b + c\cdot d)\bmod 7$  & 0/3 & 3/3 & 3/3 & 3/3 & $\sim$10\% \\
$(a\cdot b + c\cdot d)\bmod 9$  & --  & 2/3 & 3/3 & 3/3 & $\sim$10--20\% \\
$((a{+}b)\cdot(c{+}d))\bmod 10$ & --  & 0/3 & 0/3 & 3/3 & $\sim$40\% \\
$(a\cdot b + c\cdot d)\bmod 10$ & --  & 0/5 & 3/5 & 5/5 & $\sim$20--40\% \\
\bottomrule
\end{tabular}
\end{center}
\end{table}

\begin{figure}[t]
\begin{center}
\includegraphics[width=0.60\linewidth]{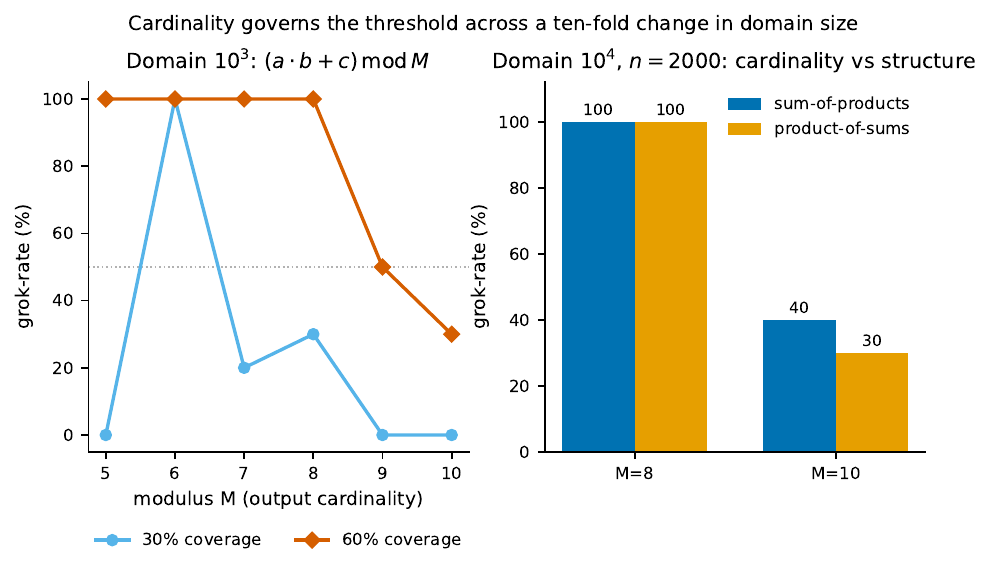}
\end{center}
\caption{Cardinality governs the coverage threshold across a ten-fold change in domain size. Left:
on the $10^3$-tuple task $(a\cdot b + c)\bmod M$, grok-rate falls with the modulus $M$ at fixed
coverage (clean at 60\%; noisier near the 30\% transition, where divisors of the digit range such
as $M=5$ are anomalous). Right: at the $n=2000$ budget on the $10^4$ domain, where the mod-10 task
is mid-transition, raising $M$ from 8 to 10 collapses the grok-rate while swapping composition
structure barely moves it; at higher budgets both moduli saturate and the gap closes
(Table~\ref{tab:struct}).}
\label{fig:cardinality}
\end{figure}

\textbf{The cardinality ordering is not an artifact of domain size, and structure is secondary.}
The four tasks above all live on the $10^4$-tuple domain, so cardinality and domain size are not
separated by them alone. We therefore ran two further controls (Figure~\ref{fig:cardinality}).
First, we repeated the coverage sweep on a three-input composite, $(a\cdot b + c)\bmod M$, whose
domain is $10^3$, an order of magnitude smaller, sweeping $M$ from 5 to 10 with ten seeds per cell.
The cardinality ordering reappears: at 60\% coverage the grok-rate is 100\% for $M\le 8$, then falls
to 50\% at $M=9$ and 30\% at $M=10$. The same monotone dependence on $M$ on a ten-fold smaller
domain shows that what sets the threshold is the number of output classes, not the absolute domain
size. The ordering is clean only well above threshold, however: at 60\% coverage it is monotone, but
near the transition at 30\% coverage it is non-monotone -- $M=5$ groks 0/10 while $M=6$ groks 10/10
-- because moduli that evenly divide the ten-digit input range impose extra residue structure on the
inputs. The $M$-dependence is thus a high-coverage regularity, not a per-seed law.

Second, to separate cardinality from structure directly, we cross two moduli with two structures
(Table~\ref{tab:struct}). The $2\times2$ is only \emph{measurable} where the higher-modulus task is
mid-transition: at $n=3000$ both moduli have saturated ($\geq$9/10 in every cell), so we read the
contrast at $n=2000$. There, swapping sum-of-products $(a\cdot b + c\cdot d)$ for product-of-sums
$((a{+}b)\cdot(c{+}d))$ at fixed modulus moves the grok-rate by at most 10 percentage points, whereas
raising the modulus from 8 to 10 at fixed structure drops it by 60--70. The cardinality effect is far
larger than the structure effect, and above threshold both vanish, exactly as a coverage-gated
transition predicts. We therefore describe the threshold as a coverage-cardinality \emph{regularity},
governed primarily by $M$ with structure a secondary, smaller-magnitude factor; the primary evidence
remains the coverage-threshold sweeps across three moduli (Table~\ref{tab:coverage} and the $10^3$
sweep above), with this $2\times2$ an illustration at a single budget.

\begin{table}[t]
\caption{Cardinality versus structure at two budgets ($10^4$ domain, ten seeds per cell). At
$n=2000$, where the mod-10 task is mid-transition, raising $M$ from 8 to 10 collapses the grok-rate
by 60--70 points while swapping structure moves it by $\leq$10; at $n=3000$ both moduli have
saturated and both gaps vanish, as a coverage-gated transition predicts. The dissociation is thus a
transition-band effect, not a budget-independent law: the primary evidence for the $M$-ordering is
the coverage-threshold sweep (Table~\ref{tab:coverage}), and this $2\times2$ only illustrates it
(the $\leq$10-point structure gap is itself within noise at ten seeds). Both mod-10 cells supersede
the 3--5-seed pilots of Table~\ref{tab:coverage} (sum-of-products 3/5$\to$4/10, product-of-sums
0/3$\to$3/10).}
\label{tab:struct}
\begin{center}
\begin{tabular}{llcc}
\toprule
budget & structure & $M=8$ & $M=10$ \\
\midrule
$n=2000$ (transition) & sum-of-products $(a\cdot b + c\cdot d)\bmod M$ & 10/10 & 4/10 \\
$n=2000$ (transition) & product-of-sums $((a{+}b)\cdot(c{+}d))\bmod M$ & 10/10 & 3/10 \\
$n=3000$ (saturated)  & sum-of-products $(a\cdot b + c\cdot d)\bmod M$ & 10/10 & 10/10 \\
$n=3000$ (saturated)  & product-of-sums $((a{+}b)\cdot(c{+}d))\bmod M$ & 10/10 & 9/10 \\
\bottomrule
\end{tabular}
\end{center}
\end{table}

This reframes the notion of a task being ``too hard'' for the model. $(a\cdot b + c\cdot d)\bmod 10$
trained at the $\sim$8\% coverage of our capability sweep groks in 0 of 10 seeds, which reads as a
hard ceiling. The same model learns the same task essentially perfectly at 40\% coverage. The
ceiling is a coverage threshold, not a capacity limit, a point we exploit in Sections~\ref{sec:wd}
and~\ref{sec:decomp}, and whose single-seed misreading we discuss in Section~\ref{sec:discussion}.

\subsection{Regularization: the Omnigrok inverted-U at 12K parameters}
\label{sec:wd}

If our grok-rate is a faithful measurement, it should respond to an intervention known to move
grokking. Weight decay is the canonical such knob: the Omnigrok account predicts that increasing
decay promotes grokking up to a point, beyond which over-regularization destroys learning. We
sweep the weight decay on $(a{+}b)\bmod 10$ at fixed 80\% coverage, ten seeds each.

\begin{figure}[t]
\begin{center}
\includegraphics[width=0.42\linewidth]{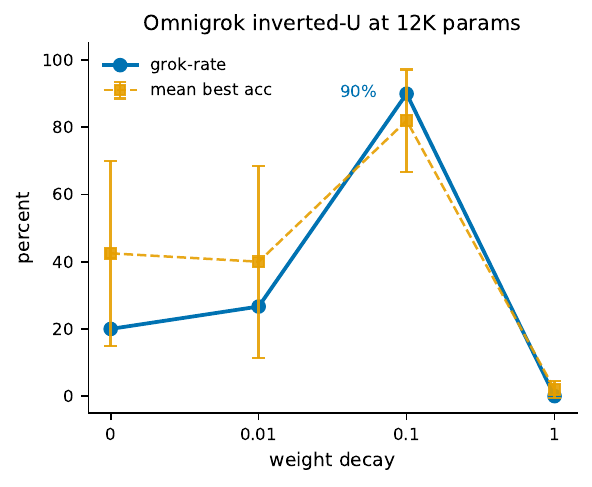}
\end{center}
\caption{Grok-rate and mean best held-out accuracy versus weight decay for $(a{+}b)\bmod 10$ (ten
seeds per point; error bars are standard deviation of best accuracy). The inverted-U, rising to
90\% at decay 0.1 and collapsing to 0\% at decay 1.0, reproduces the Omnigrok dependence at 12K
parameters and serves as a positive control for the rate measurement.}
\label{fig:wd}
\end{figure}

The result is a clean inverted-U (Table~\ref{tab:wd}, Figure~\ref{fig:wd}). Grok-rate is 20\% at
decay 0, 27\% at 0.01, \textbf{90\% at 0.1}, and 0\% at 1.0, with mean best held-out accuracy
tracking it (0.42, 0.40, 0.82, 0.02). The model reproduces the qualitative Omnigrok dependence
(decay promotes grokking, then over-regularization collapses it) at a scale two to three orders of
magnitude below the original. Two consequences follow. First, our rate measurement demonstrably
responds to a real, directional intervention, which licenses interpreting a \emph{null} effect
elsewhere (Section~\ref{sec:knifeedge}) as genuinely null rather than as measurement
insensitivity. Second, the grok-rate of a task is not an intrinsic constant: the same
$(a{+}b)\bmod 10$ groks 2--3 times in 10 at the decay we used for the capability sweep and 9 times
in 10 at decay 0.1. Any statement of the form ``this model groks task $X$ with probability $p$'' is
implicitly conditioned on the regularization.

\begin{table}[t]
\caption{Weight-decay sweep on $(a{+}b)\bmod 10$ (80\% coverage, ten seeds each). Grok-rate
follows an inverted-U, the Omnigrok signature, reproduced here at 12K parameters.}
\label{tab:wd}
\begin{center}
\begin{tabular}{lcccc}
\toprule
weight decay & 0 & 0.01 & 0.1 & 1.0 \\
\midrule
grok-rate          & 20\% & 27\% & \textbf{90\%} & 0\% \\
mean best accuracy & 0.42 & 0.40 & 0.82 & 0.02 \\
\bottomrule
\end{tabular}
\end{center}
\end{table}

\subsection{Grokking sits on a numerical knife-edge}
\label{sec:knifeedge}

We now vary the floating-point environment while holding the mathematics fixed. The cleanest such
variable is the CPU thread count. Changing the number of threads changes only the \emph{order} in
which floating-point reductions are summed; because floating-point addition is not associative
\citep{goldberg1991floating}, this perturbs the result at the level of rounding while leaving the
intended computation unchanged. We train $(a{+}b)\bmod 10$ at decay 0.01 with the thread count set
to 1, 4, and 16, pairing by seed across 300 seeds.

\begin{figure}[t]
\begin{center}
\includegraphics[width=0.64\linewidth]{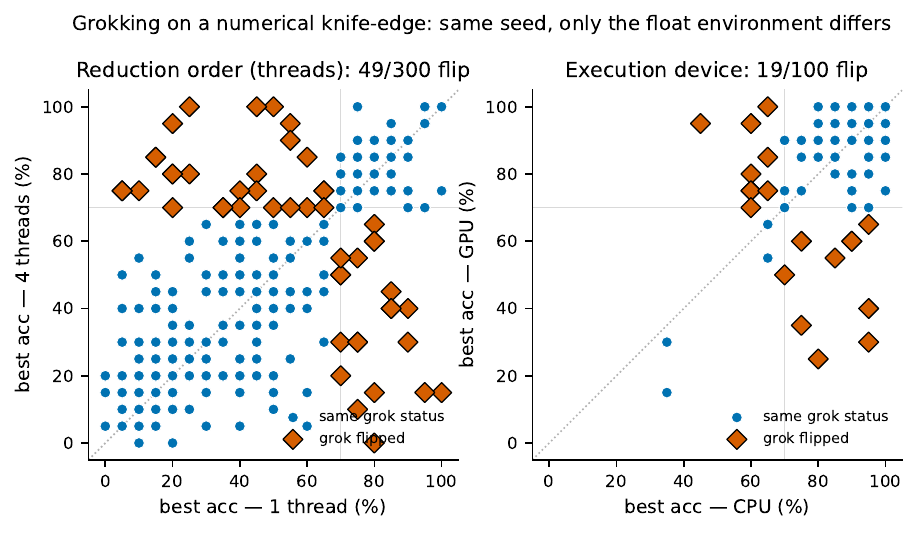}
\end{center}
\caption{Paired per-seed best accuracy under two numerical perturbations: reduction order (1 versus
4 CPU threads, 300 seeds; left) and execution device (CPU versus GPU, TF32 disabled, 100 seeds;
right). Each point is one seed; the only change between axes is the floating-point environment. Blue
points keep their grok status; orange diamonds cross the 0.70 grok threshold (flip). Flips occur at
49/300 (thread) and 19/100 (device) but straddle the diagonal near-symmetrically (exact McNemar
$p=0.39$ and $p=1.0$), so the aggregate grok-rate shows no detectable change.}
\label{fig:knifeedge}
\end{figure}

Two facts stand out (Figure~\ref{fig:knifeedge}, left). First, thread counts 4 and 16 give
bit-identical results on every seed: at this operator size the BLAS reduction tree saturates by
four threads, so 4 and 16 are the \emph{same} numerical environment. We verified this directly by
training to completion at 4 and 16 threads on a shared seed and comparing the final weights: every
parameter matches to the bit, maximum absolute difference exactly $0$. The clean perturbation is
therefore 1 versus 4. Second, this perturbation, a pure change of reduction order, flips the grok
outcome of \textbf{49 of 300 seeds (16\%)}: e.g.\ seed 57 falls from 0.80 to 0.00 (loses grokking)
and seed 132 rises from 0.25 to 1.00 (gains it). The mean absolute change in best accuracy is 0.17
and the maximum is 0.85. Yet the \emph{aggregate} grok-rate is nearly unchanged: 74/300 at one
thread versus 81/300 at four, with the 49 discordant seeds split 21 versus 28. At this seed budget
the test is well powered, and it finds \emph{no aggregate bias}: an exact two-sided McNemar test
gives $p=0.39$, and a Newcombe paired-difference 95\% confidence interval on the rate spans only
$[-6.9, +2.2]$ percentage points---ruling out any systematic shift larger than a few points. The
perturbation flips \emph{which} seeds generalize, not \emph{how many}.

To check that this is a property of grokking rather than of one particular numerical knob, we
replicate it on a second, unrelated perturbation: CPU versus GPU (Figure~\ref{fig:knifeedge},
right). We disable TF32 so the comparison is a genuine change of accumulation order at full
float32 precision, not a change of precision. Running $(a{+}b)\bmod 10$ at decay 0.1 paired by seed
across 100 seeds, the two devices use entirely different kernels and accumulation orders, yet the
grok-rate is nearly identical (86/100 on CPU, 85/100 on GPU) while 19 of 100 seeds flip (exact
McNemar $p=1.0$, Newcombe paired-difference 95\% CI $[-7.8, +9.8]$ points).
Two distinct perturbations of the floating-point environment, one (the device change) subsuming the
other (reduction order), produce a comparable signature: both flip a sixth to a fifth of same-seed
outcomes with no detectable shift in the aggregate.

Two features argue against dismissing these flips as mere proximity to a 50\% decision boundary.
First, every flip is a \emph{same-seed} event: paired runs share the data split, batch order, and
(deterministically loaded) initial weights, so only the floating-point reduction changes -- the
variance is numerical, not a reseed. Second, the two
perturbations sit at very different operating points -- the thread comparison at decay 0.01 has an
aggregate grok-rate near 25\%, the device comparison at decay 0.1 near 85\% -- yet both flip
16--19\% of seeds, so the flip rate is not an artifact of poising the aggregate exactly at 50\%. We do
not, however, claim the two flip \emph{counts} are identical across thresholds: the thread count runs
45--62/300 for $\tau\in\{0.6,0.7,0.8\}$, whereas the device count ranges 8--23/100 over the same
window (19/100 at the $\tau=0.7$ convention we report). What replicates is the qualitative signature --
a minority of same-seed flips with no detectable aggregate bias -- not a precise rate.

One reading is that the generalizing solution is reachable but sits behind a boundary fine enough
that sub-ULP differences in how a sum is accumulated decide, for a non-trivial fraction of seeds,
whether optimization crosses it. Across our 300 (thread) and 100 (device) paired seeds this resamples \emph{which} seeds generalize
without a detectable shift in \emph{how many}, unlike the directional Softmax Collapse of
\citet{prieto2025edge}; either way, the numerical environment is an uncontrolled variable that any
single run silently fixes.

\subsection{Mechanism: generalization as periodicity}
\label{sec:mechanism}

What distinguishes a generalizing solution from a memorizing one inside the weights? We train 42
add-specialists per modulus on $(a{+}b)\bmod M$ (the seed lottery supplies both grokkers and
memorizers) and correlate each model's held-out accuracy with mechanistic read-outs of its weights
and input-output map, across the full continuum rather than as a class comparison, sidestepping the
grokker/memorizer imbalance.

\begin{figure}[t]
\begin{center}
\includegraphics[width=0.42\linewidth]{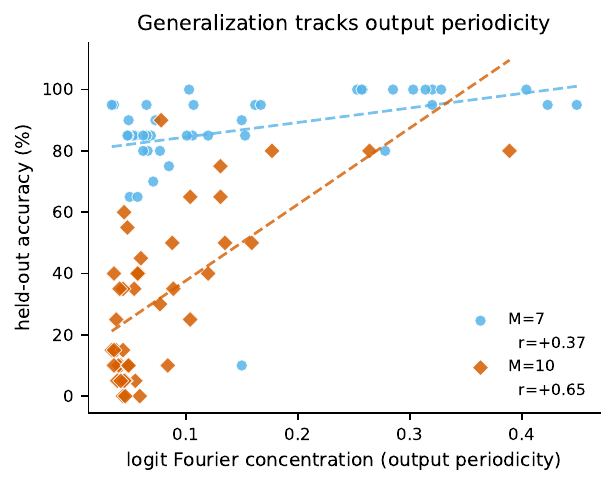}
\end{center}
\caption{Held-out accuracy versus logit-Fourier concentration (output periodicity) for 42
add-specialists per modulus. The positive correlation ($r=+0.37$ at $M=7$, $r=+0.65$ at $M=10$;
dashed fits) shows that generalizing solutions compute a more periodic input-output map.}
\label{fig:mechanism}
\end{figure}

The clearest signal is \emph{output periodicity} (Figure~\ref{fig:mechanism}). We take the logit of
the true answer token across the entire $(a,b)$ grid, compute its two-dimensional discrete Fourier
transform, and measure how concentrated the spectrum is on a single frequency. This concentration
correlates positively with generalization in both moduli: $r=+0.37$ at $M=7$ (Fisher-$z$ 95\% CI
$[+0.08,+0.61]$, Spearman $\rho=+0.55$) and $r=+0.65$ at $M=10$ (95\% CI $[+0.43,+0.80]$,
$\rho=+0.59$), with the grokker and memorizer group means separating cleanly (0.21 versus 0.06 at
$M=10$). Both intervals exclude zero and the rank correlations corroborate, though the $M=7$
interval reaches nearly to zero. Two caveats temper this. The class split is imbalanced -- 39
grokkers to 3 memorizers at $M=7$ and 5 to 37 at $M=10$ -- so each Pearson $r$ is anchored by a
small minority class and the two-cluster structure inflates $|r|$ relative to a within-cluster
trend; we therefore lead with the group-means separation and the rank correlation, and read the
$M=7$ value as no more than suggestive. More fundamentally, because the read-out is the periodicity of the
\emph{true-answer} logit over a grid generated by a periodic rule, a model that answers correctly is
periodic on that grid almost by construction; the correlation is thus partly definitional, and we
report it as a consistency check on the periodicity framing rather than as an independent causal
probe. Generalizing solutions compute a periodic, modular input-output map; memorizers compute an
aperiodic lookup -- the mechanistic counterpart of the behavioral memorization signature in
Section~\ref{sec:decomp}. The genuinely non-circular result is the embedding negative below.

We also report a clear negative result. The textbook mechanism of \citet{nanda2023progress} places the
digit embeddings on a Fourier circle, so one expects the embedding spectrum to concentrate at a few
frequencies and the embedding matrix to be low-rank. We do not see this: the per-dimension
embedding Fourier concentration is essentially uncorrelated with generalization at both moduli, and
the embedding effective rank correlates negatively at $M=10$ ($r=-0.55$, generalizers lower-rank)
but not at $M=7$ ($r\approx-0.05$, where there are only seven embedding points and little grok
variance). At hidden size 16 the model does
not allocate clean Fourier features to its embeddings; the periodicity that predicts generalization
lives in the full circuit's output map, not in the raw embedding geometry. We thus report a weaker,
but well-supported, version of the Fourier-multiplication picture.

\subsection{Decomposition as data efficiency}
\label{sec:decomp}

Finally we revisit the appeal of multi-agent decomposition. A composite such as
$(a\cdot b + c\cdot d)\bmod 10$ can be solved by a single model (a monolith) or by a pipeline: a
multiplication specialist computes $a\cdot b$ and $c\cdot d$, and an addition specialist combines
them. The natural claim (that the monolith cannot do the task and decomposition rescues it) is, by
Section~\ref{sec:coverage}, false: the monolith groks the composite at sufficient coverage and
regularization. The real question is whether decomposition helps \emph{at a matched data budget
and matched regularization}.

\begin{figure}[t]
\begin{center}
\includegraphics[width=0.64\linewidth]{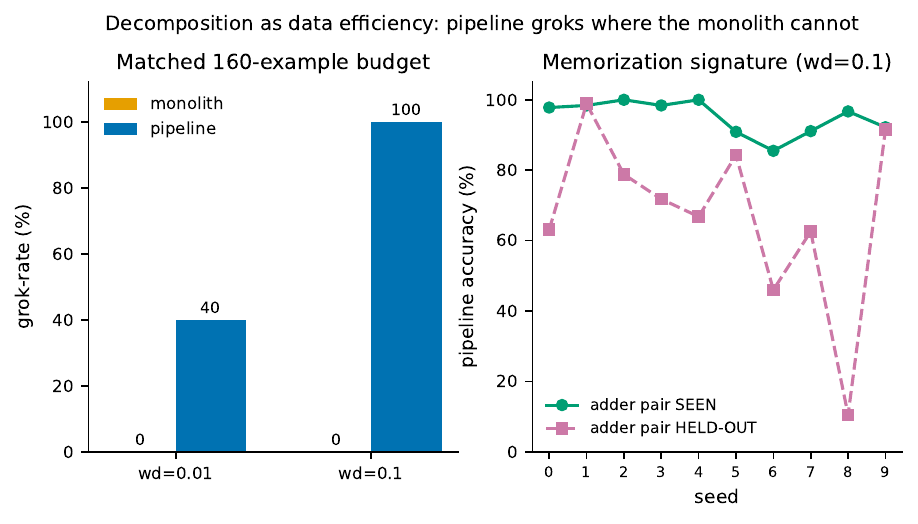}
\end{center}
\caption{Left: grok-rate of monolith versus pipeline at a matched 160-example budget; the pipeline
groks (40\% at decay 0.01, 100\% at decay 0.1) where the monolith never does (0\%). Right: per-seed
pipeline accuracy on intermediate sum-pairs the adder saw in training (SEEN) versus unseen pairs
(HELD-OUT) at decay 0.1; the large SEEN$\gg$HELD-OUT gap is the memorization signature, and it
closes for the seeds whose adder genuinely generalized (1, 9).}
\label{fig:decomp}
\end{figure}

It does, decisively (Table~\ref{tab:decomp}, Figure~\ref{fig:decomp}). We give the monolith 160
composite training examples and give the pipeline the same total budget (80 multiplication examples
and 80 addition examples), and evaluate both on the same held-out 1{,}000 composite tuples. At
decay 0.01 the monolith groks 0/10 (mean best 0.20) while the pipeline reaches 0.70 and groks
4/10; at decay 0.1 the monolith still groks 0/10 (mean 0.21) while the pipeline reaches 0.89 and
groks \textbf{10/10}. The pipeline beats the monolith on every one of the ten seeds at both decays,
by 50 and 68 percentage points of mean accuracy respectively (and it lifts the grok-rate from 0/10
to 4/10 and 10/10). Combined with the observation that the monolith needs roughly 800--2{,}000
composite examples to grok at all (2--4/10), the effect is roughly an order-of-magnitude improvement
in data efficiency, plus a higher grok-rate.

\begin{table}[t]
\caption{Pipeline versus monolith on $(a\cdot b + c\cdot d)\bmod 10$ at a matched 160-example
budget, ten seeds. The \emph{scratchpad} monolith adds the pipeline's decomposed intermediate
supervision at the same sparse coverage; it does not grok either, isolating coverage rather than
supervision as the driver. The pipeline groks where both monoliths never do, and the gain widens at
the grok-friendly decay 0.1.}
\label{tab:decomp}
\begin{center}
\begin{tabular}{lcccccc}
\toprule
 & \multicolumn{2}{c}{monolith (direct)} & \multicolumn{2}{c}{monolith (scratchpad)} & \multicolumn{2}{c}{pipeline} \\
\cmidrule(lr){2-3}\cmidrule(lr){4-5}\cmidrule(lr){6-7}
weight decay & grok-rate & mean best & grok-rate & mean best & grok-rate & mean acc \\
\midrule
0.01 & 0/10 & 0.20 & 0/10 & 0.27 & 4/10          & 0.70 \\
0.1  & 0/10 & 0.21 & 0/10 & 0.31 & \textbf{10/10} & 0.89 \\
\bottomrule
\end{tabular}
\end{center}
\end{table}

The mechanism is exactly the coverage story of Section~\ref{sec:coverage} read in reverse. The
160-example composite budget covers only 1.6\% of the sparse $10^4$ composite domain, far below
threshold, but the same examples cover 80\% of each dense $10^2$ sub-domain, well above it.
Decomposition wins by spending the data where coverage is cheap. Reading the pipeline's internals
confirms a memorization mechanism: the addition specialist reaches 100\% training accuracy, and the
pipeline is accurate on the intermediate sum-pairs the adder saw in training and much weaker on
unseen pairs (seed 8 scores 0.97 on seen versus 0.10 on held-out), the gap vanishing precisely for
the seeds whose adder genuinely generalized (seed 1, 0.98 versus 0.99) -- a memorization signature,
not an artifact. Decomposition, here, turns a generalization problem into a tractable memorization
problem by making each sub-domain dense.

A confound qualifies the matched-budget framing, and we can separate it. The pipeline's examples
carry not only the same \emph{count} of labels but more \emph{decomposed} supervision: the
specialists see the intermediate products and sum as direct targets, whereas the monolith must infer
them from the composite label alone. Coverage and supervision density are thus confounded by
construction. To isolate them we add a third arm (Table~\ref{tab:decomp}): a \emph{scratchpad}
monolith trained on the same 160 tuples, as a single model, but to emit the two products and their
sum as an explicit chain $P,Q,R=(a\cdot b)\bmod 10,\,(c\cdot d)\bmod 10,\,(P{+}Q)\bmod 10$ --
exactly the pipeline's decomposed supervision, with coverage held at the sparse composite level. It
does not close the gap: the scratchpad monolith groks 0/10 at both decays (mean best 0.27 and 0.31,
barely above the direct monolith's 0.20 and 0.21), and emits the correct intermediate pair on
held-out tuples only 31\% of the time -- at 1.6\% coverage it cannot even reconstruct the
intermediates it is supervised on. Decomposed supervision alone therefore does not rescue the
pipeline; the operative factor is the dense sub-domain coverage that lets each specialist generalize,
and the confound resolves in favor of coverage over supervision density.

\section{Discussion}
\label{sec:discussion}

\textbf{Single runs lie in this regime, systematically.} The throughline of our results is that the
grokking transition in a small model is conditional on coverage (Section~\ref{sec:coverage}),
regularization (Section~\ref{sec:wd}), and the numerical environment (Section~\ref{sec:knifeedge}),
and that any single run silently fixes all three. We encountered this not as an abstraction but as
three concrete near-misses, each a clean story on one seed: a ``capability wall'' (0/10) that groks
once the decay is raised; a thread-count change that seemed to move the grok-rate 95\%$\to$55\%; and
a GPU that seemed to suppress grokking (45\% versus 95\%). Paired across seeds all three rates are
flat -- only individual outcomes move, the apparent effects seed confounds dressed as mechanisms.
The practical consequence is a short protocol: report grok-\emph{rates} over many seeds; pin and
report the thread count and device; pair by seed when comparing conditions; and treat a dramatic
single-run effect as a hypothesis to be killed by control. We suspect this is a hazard wherever a
phase transition is measured near its threshold, not a quirk of our model.

\textbf{Why this matters beyond a 12K model.} The specific knife-edge we characterize belongs to this small, tractable regime, and we do not claim the same sub-ULP sensitivity persists unchanged at scale. What transfers is the methodological hazard. Sharp, threshold-like transitions are increasingly invoked in claims about large models---grokking, emergent abilities, capability ``walls''---and such transitions are settings where an uncontrolled seed or numerical environment could, in principle, help decide a reported outcome. Our tractable setting lets us show, against a direct reading of the weights, that two innocuous-looking choices (how many threads a reduction uses; CPU versus GPU) flip a sixth of same-seed outcomes with no aggregate signal, and that three clean single-run narratives in our own data dissolve under paired, multi-seed control. The prescription that follows is scale-independent: report rates over many seeds, pin and report the numerical environment, pair by seed, and treat a dramatic single-run transition as a hypothesis to be refuted rather than a result. As the phenomena that most interest the field grow sharper and are increasingly read from expensive single runs, the cost of neglecting this rises. We offer the fully-legible regime less as a claim about 12K-parameter models than as a controlled testbed for a reproducibility discipline that studying phase transitions at any scale requires.

\textbf{What fragility implies.} The reduction-order result constrains the loss landscape: the
generalizing basin is reachable but shallow enough that sub-ULP accumulation differences decide
entry for a sixth of seeds. This fits the Omnigrok picture of a slow drift to a distant, narrow
region of weight space and the numerical-edge picture of \citet{prieto2025edge}; our addition is
that the edge is sharp enough for \emph{benign} numerical noise, not only catastrophic instability,
to have macroscopic consequences -- so the right unit for theories of grokking is a distribution
over runs, not a representative one.

\textbf{Limitations.} Our model is a single toy architecture; the tasks are modular arithmetic with
$M \leq 10$, chosen for single-token answers; the numerics we characterize are CPU-specific BLAS
reductions and a single GPU. The grok threshold of 0.70 is a convention; because a discontinuous metric can manufacture an
apparent transition \citep{schaeffer2023mirage}, we re-tabulated every grok-rate at thresholds
0.60, 0.70, and 0.80. The inverted-U, the thread flip count (45--62/300 across the three thresholds), and the
decomposition gap are qualitatively unchanged (the weight-decay grok-rates read 20/27/90/0\% at
threshold 0.70 and 20/17/70/0\% at 0.80); the device flip count is more threshold-sensitive
(8--23/100), but its McNemar test is non-significant at every threshold, so no conclusion is
load-bearing on the cutoff. The cardinality regularity should
nonetheless be tested even more densely, at larger $M$, and in input bases other than the base-10
digits used throughout, before it is read as more than an in-setting regularity. The mechanistic read-out establishes correlation between periodicity and generalization, not
the full causal circuit that \citet{nanda2023progress} recover at larger width. Whether the
numerical knife-edge persists at scale, where averaging over many parameters may damp sub-ULP
effects, is an open and important question.

\section{Conclusion}
\label{sec:conclusion}

In a model small enough to enumerate end to end, grokking is not a property a task either has or
lacks. It is a conditional, fragile phase transition: gated by how much of the input domain is
covered, promoted and then destroyed by weight decay, and balanced so finely that the order in
which floating-point numbers are added flips the outcome for a sixth of random seeds without
detectably changing how often it happens overall. Generalizing solutions have a more periodic
output map than memorizing ones, though we read this as a consistency check; the genuinely
independent mechanistic result is that the embeddings do not form the textbook Fourier circle.
Decomposition helps primarily by making coverage cheap. Across all of these, the discipline that
produced reliable findings was the same: measure a rate over many seeds inside a fixed numerical
environment, and distrust any clean story a single run tells. We offer the fully-tractable setting
as a testbed where that discipline can be checked against a direct reading of the weights and
input-output map; whether these findings carry to larger scales we leave open, claiming only the
fully-legible regime in which grokking can be dissected rather than merely observed.

\subsubsection*{Broader Impact Statement}
This work studies the training dynamics of a publicly released toy model on synthetic arithmetic,
with no human subjects or sensitive data. Its main implication is methodological: claims about
grokking and similar phase transitions should rest on multi-seed rates under a controlled numerical
environment, not single runs. We foresee no negative societal impact specific to this study.

\subsubsection*{Reproducibility Statement}
All experiments use the public Glint-Research/Glimmer-1-Base checkpoint; all scripts, per-seed
records, and this paper's source are released in a public repository,\footnote{\url{https://github.com/otanl/grokking-microscope}, archived at \url{https://doi.org/10.5281/zenodo.21221306}} and every
reported number -- including the paired statistics (exact
McNemar, Newcombe, Fisher-$z$, Spearman) -- is reproducible from them. The numerical environment is
recorded per run, as it is load-bearing: Python 3.12.9, PyTorch 2.12.1+cu126, Transformers 5.12.1,
all sweeps on CPU in float32, with the device-control leg on a single NVIDIA RTX 3090 Ti (TF32 disabled---the device comparison is a pure change of accumulation order at full float32; the CPU BLAS backend is Intel MKL on an AMD Zen~2 CPU; the thread count is pinned and reported per experiment).

\bibliography{main}
\bibliographystyle{tmlr}

\end{document}